\documentclass[%
 reprint,
superscriptaddress,
 amsmath,amssymb,
 aps,
pra,
]{revtex4-2}
\usepackage[utf8]{inputenc}
\usepackage{graphicx}
\usepackage{dcolumn}
\usepackage{bm}
\usepackage{float}
\usepackage{cancel}
\usepackage{bm}
\usepackage{bbm}
\usepackage{amssymb}
\usepackage{amsmath}
\usepackage{amsthm}
\usepackage{color}
\usepackage{xcolor}
\usepackage{booktabs}
\usepackage{hyperref}
\usepackage{siunitx}
\usepackage{lipsum}

\usepackage{amsmath,amsfonts,bm}

\def\eqref#1{equation~\ref{#1}}

\def\1{\bm{1}}

\def\vmu{{\bm{\mu}}}
\def\vtheta{{\bm{\theta}}}
\def\vlambda{{\bm{\lambda}}}
\def\vvarphi{{\bm{\varphi}}}
\def\vPhi{{\bm{\Phi}}}
\def\vphi{{\bm{\phi}}}
\def\vxi{{\bm{\xi}}}

\def\ve{{\bm{e}}}

\def\vp{{\bm{p}}}

\def\vu{{\bm{u}}}

\def\vx{{\bm{x}}}
\def\vy{{\bm{y}}}
\def\vz{{\bm{z}}}

\def\mA{{\bm{A}}}
\def\mB{{\bm{B}}}
\def\mC{{\bm{C}}}
\def\mD{{\bm{D}}}

\def\mG{{\bm{G}}}

\def\mI{{\bm{I}}}

\def\mK{{\bm{K}}}

\def\mO{{\bm{O}}}

\def\mQ{{\bm{Q}}}

\def\mU{{\bm{U}}}
\def\mV{{\bm{V}}}
\def\mW{{\bm{W}}}

\def\mZ{{\bm{Z}}}

\def\mLambda{{\bm{\Lambda}}}

\def\mDelta{{\bm{\Delta}}}

\DeclareMathAlphabet{\mathsfit}{\encodingdefault}{\sfdefault}{m}{sl}
\SetMathAlphabet{\mathsfit}{bold}{\encodingdefault}{\sfdefault}{bx}{n}

\DeclareMathOperator{\Tr}{Tr}

\graphicspath{Figures}

\begin{document}

\title{When Does Depth Matter For In-Context Learning? Adaptive Inference in Deep Transformers}

\author{Ravin Raj}
\author{Gautam Reddy}
\email{greddy@princeton.edu}
\affiliation{
Department of Physics, Princeton University, Princeton, NJ 08544, USA
}

\date{\today}

\begin{abstract}
Transformers perform computations through many successive attention and feedforward blocks, allowing them to learn complex correlations between a large collection of coupled variables. When does stacking successive attention-feedforward computations over many layers provide a computational advantage over a single transformer block? We address this question by examining in-context learning in generalized linear attention transformers. We first introduce a general theory of distributed inference in such transformers, subject to constraints on communication and depth. We show that such systems can exploit internal representations (`function vectors') to infer a latent context variable at increasingly finer scales over its layers. For an in-context linear regression task, the theory predicts that while one-layer transformers without feedforward blocks are optimal for Gaussian priors over the context variable, multi-layer transformers are superior for non-Gaussian, tree-like priors. Trained linear attention transformers reproduce quantitative predictions from the theory. Using causal key-patching experiments, we verify that function vectors in intermediate layers mediate adaptive routing of information. Our results suggest that depth and feedforward blocks enable transformers to implement adaptive inference, and this is advantageous when the distribution over latent variables has hierarchical structure.
\end{abstract}

\maketitle


\section{Introduction}

Often, one is interested in describing the joint distribution of a large collection of coupled random variables, $P(\vz_1,\vz_2,\dots, \vz_N)$. Many classical techniques have been developed to estimate this distribution from data, often requiring simplifying ans\"atze that allow for efficient inference~\citep{Bishop2006}. These techniques have been largely superceded in recent years by learning methods that optimize a flexible architecture using large datasets~\citep{ortega2019meta, muller2021transformers, genewein2026algorithmic}. The transformer architecture has emerged as one such general-purpose substrate~\citep{vaswani_attention_2017}
, and now forms the basis of foundation models in many domains beyond language, including vision~\citep{dosovitskiy_image_2020,radford_learning_2021,kirillov_segment_2023}, speech~\citep{baevski_wav2vec_2020,hsu_hubert_2021,radford_robust_2023}, materials design~\citep{qu_leveraging_2024} and molecule generation~\citep{irwin_chemformer_2022, bagal_molgpt_2022, chandra_transformer-based_2023, wei_probabilistic_2023}. 

The flexibility of a transformer arises from two operations that are repeated across layers. The first is a global interaction, mediated by self-attention, in which each vector (or `token') in the input sequence extracts information from other tokens through learned interaction maps. The second is a local nonlinear computation, usually implemented as a multi-layer perceptron (MLP), which acts independently and identically on each vector. Once its parameters are optimized, a transformer can adapt its predictions to a novel sequence or dataset in a single forward pass, without further updates to its parameters. This phenomenon, called in-context learning (ICL), suggests that the transformer has learned a generic procedure for inferring structure from context~\citep{brown_language_2020, olsson_-context_2022}. In-context learning has served as a paradigm for probing the internal mechanisms of transformers~\citep{xie_explanation_2022,garg_what_2022,akyurek_what_2022, bai2023transformers}. We will use this paradigm for a similar purpose. 

While in large language models, inference is typically performed through many successive layers of attention and feedforward (MLP) blocks, mechanistic understanding of ICL is restricted to shallow models and typically without feedforward blocks (reviewed below). This leaves us with the following question: \textbf{when does depth provide a computational advantage?} Our hope is that addressing this question in controlled, simplified scenarios will help explain the remarkable success of deep models on practically relevant tasks. We approach this question for a class of architectures we call \emph{generalized linear attention transformers}. Predictions from theory are tested on an in-context linear regression task using one derived variant from this class. 

\subsection{Related Work}

\paragraph{Algorithmic and mechanistic accounts of ICL.}
A plethora of work has attempted to characterize the computation implemented by transformers during ICL. This work has identified two classes of motifs that implement ICL: (1) attention circuits that implement specific algorithmic operations, such as copying~\citep{olsson_-context_2022, reddy2023mechanistic, bietti2023birth} or gradient descent~\citep{oswald_transformers_2023, ahn_transformers_2023, park_algorithmic_2024, zhang2024trained}, and (2) by forming \emph{function vectors}~\citep{todd_function_2023,hendel_context_2023,hojel_finding_2024, yin2025attention}, which are internal representations of the function or task the model deems is being solved. While recent theoretical efforts have shed light on the former class of mechanisms~\citep{reddy2023mechanistic, lu2025asymptotic, ahn_transformers_2023, oswald_transformers_2023, zhang2024trained, zhang2025training, he2025context, giannou2024well}, the transformer circuits that enable the latter have remained unclear. First identified in large language models~\citep{hendel_context_2023,todd_function_2023}, function (or task) vectors are consistently found in the middle layers of billion-parameter models of different families, suggesting a common underlying principle that favors this organization. Recent empirical work~\citep{gibson_distinct_2026} finds that, in contrast to pure attention-based circuits, leveraging function vectors necessarily entails a multi-layer interaction between the attention and MLP blocks.
The current work builds on function-vector-based ICL. 

\paragraph{The role of depth in ICL.}
Recent work has characterized benefits of depth arising from iterative computation within in-context learning. \citet{bordelon_theory_2026} show that deep linear attention transformers can help in-context linear regression when the input covariance varies across contexts, thus preventing effective preconditioning of the data before applying gradient descent. Depth can help by applying multiple steps of unconditioned gradient descent.  \citet{sun_role_2026} show that MLPs can implement nonlinear regression by constructing polynomial-kernel features, so that in-context gradient descent operates on nonlinear features. Here, we propose a general theory that highlights how the structure of latent context variables influences the advantage offered by depth. We show that MLPs play other computational roles beyond nonlinear feature construction. In particular, MLPs use intermediate task representations to route which information is exchanged in subsequent attention layers and further act as decoders that apply the inferred function on a novel query. 

\paragraph{Adaptive inference.}
Adaptive inference is rooted in statistics and has its origins in the theory of sequential experimental design~\citep{chernoff1992sequential,chaloner1995bayesian}. The problem involves choosing what to measure next based on the outcomes of previous measurements, which can be significantly more efficient when multiple rounds of measurements are possible. Optimal design often involves acquiring information about an underlying latent variable. \citet{braun_infogreedy_2015} developed a framework for adaptive sensing by sequentially choosing measurements through maximizing information gain conditioned on previous observations. Information-based search strategies, such as infotaxis~\citep{vergassola2007infotaxis}, rely on a similar principle. Our work builds on this principle, showing that a transformer's forward pass can approximate an adaptive inference procedure through successive interactions between self-attention and feedforward blocks.

\subsection{Summary of Contributions}
Here, we propose that depth enables a transformer to perform adaptive inference. Particularly, a shallow model must choose all of the information it extracts from the context simultaneously, while a deep model can refine later measurements based on information acquired at earlier layers. In this sequential layerwise approach, the accumulated function vector acts as a state variable, feedforward (MLP) blocks determine which statistic should be extracted next, and attention pools that statistic across the context. Depth can therefore enable an adaptive search over the hypothesis space, focusing computation on hypotheses which remain plausible given earlier observations. 
In summary, our main contributions are:
\begin{itemize}
    \item \textbf{A framework for isolating the role of depth in ICL}. We formulate in-context inference under a fixed communication budget, allowing shallow and deep models to be compared on equal footing while controlling for the total amount of information extracted from the context.
    \item \textbf{A theory of when depth provides an advantage}. We frame layerwise inference as a sequential decision problem and show that depth is useful only when information acquired at one layer changes the choice of information extracted next.
    \item \textbf{A mechanistic realization in trained transformer models}. Deep transformers learn the predicted adaptive inference strategy, and causal key-patching experiments show that intermediate context representations control the routing decisions made at subsequent layers.
\end{itemize}

\section{In-context learning in deep transformers}
\label{sec:problem_setup}
\noindent \textbf{In-context learning.} We wish to learn the joint distribution of a collection of $N$ random variables (`tokens'), $\mZ_N = \{\vz_1, \vz_2, \ldots, \vz_N\}$. We consider a standard in-context learning paradigm~\citep{garg_what_2022, ahn_transformers_2023, zhang2024trained, lu2025asymptotic}. Specifically, the tokens are independent conditional on a latent context variable $\vtheta$ (drawn from $\rho_0(\vtheta)$), which allows us to write
\begin{align}P(\{\vz_i\}) = \int d\vtheta \rho_0(\vtheta) \prod_{i} p(\vz_i|\vtheta).
\end{align}
The model is trained to predict a masked-out component in a separate query token using the context $\mZ_N$. Specifically, each token is split into two components, $\vz = (\vx, \vy)$, where $p(\vz |\vtheta) = q(\vx | \vtheta) \pi(\vy | \vx, \vtheta)$. Given $\mZ_N$ and an independently drawn query token $\vx$ from the same $\vtheta$, the model is trained to construct a predictive distribution $\hat{\pi}(\vy|\vx, \mZ_N) \approx \pi(\vy | \vx, \vtheta)$ by optimizing a prediction loss $\langle \mathcal{L}(\hat{\pi}) \rangle$. $\mathcal{L}$ is typically a cross-entropy loss or a mean-squared-error loss, and the expectation is over samples of $\vtheta \sim \rho_0$, a new set of contextual tokens $\mZ_N$ given $\vtheta$, and a new query $\vx \sim q(\cdot|\vtheta)$.
\\

\begin{figure}[t]
    \begin{center}
    \includegraphics[width=\linewidth]{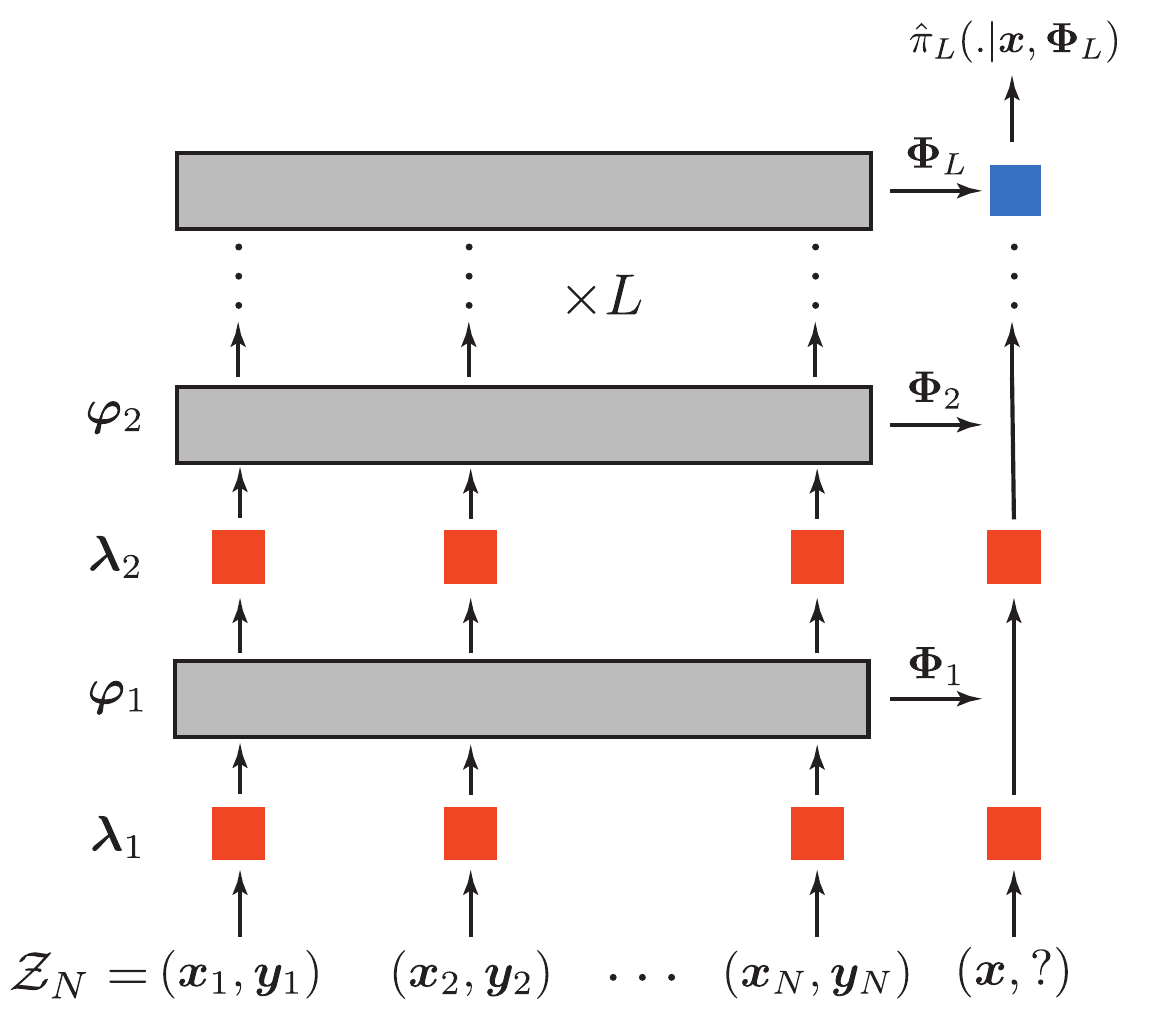}
    \end{center}
    \caption{\textbf{Model schematic.} Information about a latent contextual variable $\vtheta$ implicit in the context $\mathcal{Z}_N$ is compressed through $L$ rounds of interaction into a function vector $\bm{\Phi}_L$, which is subsequently used by a query for prediction (through a decoder; blue square). At every layer $\ell + 1$, the local nonlinearity (red square) uses $(\vx_i, \vy_i)$ and the intermediate function vector $\bm{\Phi}_{\ell}$ to construct an embedding function $\vlambda_{\ell+1}(\vx_i, \vy_i, \bm{\Phi}_{\ell})$. The embedding of the $N$ contextual tokens is pooled (gray rectangle) to form a $d$-dimensional statistic $\vvarphi_{\ell+1}$. This pooled statistic is concatenated: $\bm{\Phi}_{\ell+1} = \bm{\Phi}_{\ell} \oplus \vvarphi_{\ell+1}$.}
    \label{fig:model_schematic}
\end{figure}

\noindent \textbf{Generalized linear attention transformer.} We consider transformer architectures that consist of $L$ layers, where each layer includes one step of pooling interactions between tokens and one step of a local nonlinear operation that processes each token separately (Figure~\ref{fig:model_schematic}). Importantly, interactions are constrained to a dimension $d$. This constraint will allow us to compare architectures of different depths under a fixed communication budget. We assume a common masking scheme where the context tokens cannot interact with the query, whereas the query $\bm{x}$ can access the shared communication channel among the context tokens. Since a pooling operation is linear in the number of tokens $N$, the computational complexity of this architecture scales linearly with $N$. Thus, this architecture can be viewed as a \emph{generalized linear attention transformer}. Indeed, we will see later that commonly used linear attention transformers are special cases with interaction dimension $d = d_K d_V$, where $d_K, d_V$ are the key and value dimensions respectively~\citep{wang2020linformer,katharopoulos2020transformers,schlag2021linear,choromanski_rethinking_2022}. 
We then show that \emph{optimal} in-context learning with this architecture involves adaptive inference of the latent context variable $\bm{\theta}$. 
\\

\noindent \textbf{One-layer model.} Let's begin with a one-layer model, $L = 1$. Since $\vtheta$ is a shared variable, the tokens should exchange information to perform inference. We assume interactions are mediated through a shared pool
\begin{equation}
    \vvarphi_1= \frac{1}{N} \sum_{i=1}^N \vlambda_1(\vz_i), \label{eq:fv1}
\end{equation}
where $\vlambda_1$ is a $d$-dimensional nonlinear embedding of each token and $\vvarphi_1$ is a globally accessible statistic that carries information about $\vtheta$.  
At optimality, the query $\vx$ can access $\vvarphi_1$ to form a posterior $\rho(\vtheta | \vx, \vvarphi_1)$ and thereby generate a predictive distribution $\hat{\pi}(\vy | \vx, \mZ_N) = \int d\vtheta \rho(\vtheta | \vx, \vvarphi_1) \pi(\vy | \vx,\vtheta)$. 
Since information about the context is shared to the query entirely through $\vvarphi_1$, we define $\vvarphi_1$ as the \emph{function vector} corresponding to the context $\mZ_N$. This definition of a function vector aligns with empirically motivated definitions of function/task vectors~\citep{todd_function_2023, hendel_context_2023, gibson_distinct_2026}: an internal representation of a model is deemed a function vector if patching in the function vector from context $\mZ_N$ into a different context $\mZ'_N$ produces a response to a novel query $\vx$ in context $\mZ'_N$ that aligns with context $\mZ_N$. In other words, a function vector is a compact summary of the information in the context for predicting the target $\vy$ for any novel query $\vx$. 
\\

\noindent \textbf{Function vectors as internal state variables.} Now, suppose $L \ge 1$. After the first $\ell$ layers, define
\begin{equation}
    \bm{\Phi}_{\ell} = \vvarphi_{1}\oplus \vvarphi_{2}\oplus \dots \oplus \vvarphi_{\ell}, \qquad \bm{\Phi}_0=\varnothing,
\end{equation}
where $\oplus$ denotes a direct sum of vectors and the function vector at layer $\ell$ is $\vvarphi_{\ell} = \sum_{i=1}^N\vlambda_{\ell}(\vz_i, \bm{\Phi}_{\ell})/N$. Crucially, $\bm{\Phi}_{\ell}$ acts as a state variable for implementing adaptive inference. To see this, consider the following heuristic argument.
Let us say we want to adaptively select the embedding at the next layer $\vlambda_{\ell+1}$ to optimize the final prediction loss after $L$ layers. A careful choice is necessary because of the communication constraint imposed by the finite dimension $d$ of $\vlambda_{\ell+1}$. One may imagine, however, that making this choice requires keeping track of the embeddings at all preceding layers, $\vlambda_{1}, \vlambda_{2}, \dots, \vlambda_{\ell}$ (which, problematically, are functions rather than statistics). However, storing this history is not necessary in principle. Specifically, since $\bm{\Phi}_{\ell}$ preserves all statistics acquired so far, and because the embedding at each layer is chosen as a function of the previous state, $\bm{\Phi}_{\ell}$ \emph{implicitly} records the choices of embeddings $\{\vlambda_{1:\ell}\}$ that produced those statistics. 
\\

\noindent \textbf{Adaptive inference.} We now examine how the state variable $\bm{\Phi}_{\ell}$ can be used to optimally choose $\vlambda_{\ell+1}$. Denote $\hat{\pi}_L(\vy|\vx,\bm{\Phi}_L) =\int d\vtheta \, \rho(\vtheta|\vx, \bm{\Phi}_L)\pi(\vy|\vx,\vtheta)$
for the predictive distribution after the final layer, and define the terminal value $V_L(\bm{\Phi}_L)= \int d\vtheta \,\rho(\vtheta | \bm{\Phi}_L) \int d\vz \,p(\vz|\vtheta)\mathcal{L}(\hat{\pi}_L)$, which is the expected loss averaged over the posterior of possible latent context variables $\vtheta$ and queries $\vx$. 

For $\ell<L$, say $\vvarphi$ is the random statistic generated by a candidate next-layer embedding $\vlambda$, and let
$P_{\vlambda}(\vvarphi|\bm{\Phi}_{\ell})$ denote its conditional distribution. $P_{\vlambda}(\vvarphi|\bm{\Phi}_{\ell})$ implicitly includes a complicated average over all possible contexts $\mZ_N$ compatible with $\bm{\Phi}_{\ell}$. By the principle of dynamic programming~\citep{Bellman1957}, the optimal value formally satisfies the equation:
\begin{equation}
    V_{\ell}(\bm{\Phi}_{\ell}) = \min_{\vlambda} \left\langle V_{\ell+1}(\bm{\Phi}_{\ell}\oplus \vvarphi) \right\rangle_{\vvarphi \sim P_{\vlambda}(\cdot | \bm{\Phi}_{\ell})}. \label{eq:bellman}
\end{equation} 
This so-called Bellman equation is central to optimal control and appears here because one has to choose interactions at an intermediate layer to minimize the prediction loss using partial information from earlier layers. While this equation is challenging to solve analytically in most cases, it nevertheless shows that the optimal embedding $\vlambda_{\ell+1}$, which is the minimizer in Eq.~\ref{eq:bellman}, should be interpreted as an experimental design strategy for adaptive inference~\citep{chaloner1995bayesian, chernoff1992sequential}. That is, the information exchanged by tokens at each layer should constrain the posterior distribution over $\bm{\theta}$ such that the model ultimately minimizes prediction loss at the final layer. Once $\vlambda_{\ell+1}$ is chosen, the model appends the corresponding statistic $\vvarphi_{\ell+1}$ to the state, so that $\bm{\Phi}_{\ell+1}=\bm{\Phi}_{\ell}\oplus \vvarphi_{\ell+1}$. After $L$ layers, the query $\vx$ uses $\bm{\Phi}_L$ to form $\hat{\pi}_L(\vy|\vx,\bm{\Phi}_L)$. 
\\

\noindent \textbf{The role of attention and feedforward blocks at optimality.} This discussion suggests that, if the transformer achieves optimality after training, then the feedforward blocks at convergence play at least two distinct computational roles. Their role depends on whether they are in intermediate layers or in the final layer: i) in the intermediate layers, they act as \emph{routers} that choose which information in the token should be conveyed to other tokens, i.e., $(\vz, \bm{\Phi}_{\ell}) \to \vlambda_{\ell+1}$, and ii) in the final layer, the MLP acts as a \emph{decoder} that processes information gathered from other tokens to generate a prediction, $(\vx, \bm{\Phi}_{L}) \to \hat{\pi}$. The attention operation mediates interactions and builds the function vector by pooling information across tokens, $\{\vlambda_{\ell}(\vz_i,\bm{\Phi}_{\ell-1})\} \to \vvarphi_{\ell}$. To achieve optimality, the discussion further implies that the residual stream should be sufficiently large to retain information about the original token $\vz_i$ and all the function vectors from previous layers. Motivated by this picture, we introduce a novel linear attention transformer architecture in Section~\ref{sec:tf_results} and discuss its relationship with commonly used architectures in Appendix~\ref{app:transformer_mapping}.

\section{In-context linear regression and the role of depth}
\label{sec:depth}
The adaptive inference framework allows us to derive a non-trivial relationship between the structure of the prior $\rho_0$ and the advantage offered by depth and MLP blocks in transformers. We examine this relationship in more detail in an in-context linear regression task~\citep{garg_what_2022}. In each context, we generate tokens such that
\begin{align}
    y = \vtheta^\top\vx + \eta&, \quad \vx \sim \mathcal{N}(\bm{0}, \mI_K), \nonumber \\
    \quad \eta \sim \mathcal{N}(0,\sigma^2)&, \quad \vtheta \sim \rho_0.
\end{align}
For a context $\mathcal{Z}_N$ drawn from $\vtheta$, the loss is $\mathcal{L} = (\vtheta^\top\vx - \hat{y}(\vx,\mathcal{Z}_N))^2$, where $\hat{y}(\vx,\mathcal{Z}_N)$ is the predicted output for query $\vx$. For the theoretical analysis, we work in the limit $N \to \infty$ and the noise variance is scaled by fixing $\sigma^2/N = \tau^2$ to retain noise in this limit. 

To derive predictions, we consider embeddings of the form $\vlambda_{\ell} = y\mW_{\ell}\vx$. This functional form for $\vlambda_{\ell}$ is motivated by the key-value outer product structure in linear attention architectures, which we will later use to test predictions. For this choice, the function vectors can be easily calculated: 
\begin{align}
    \vvarphi_{\ell} = \mW_{\ell}(\vtheta + \vxi), \quad \vxi\sim \mathcal{N}(\bm{0},\tau^2 \mI_K),\label{eq:phi_linear}
\end{align}
where $\vxi$ is due to the noise $\eta$ and is fixed for all layers. 
One can interpret the rows of $\mW_{\ell}$ as forming the subspace along which $\vtheta$ is measured and the function vector $\vvarphi_{\ell}$ as the outcome of the measurement. From the definition of $\bm{\Phi}_{\ell}$, we have $\bm{\Phi}_{\ell} = \mA_{\ell}(\vtheta + \vxi)$, where $\mA_{\ell} = \mW_1\oplus \mW_2\oplus \dots \oplus \mW_{\ell}$. Recall that for an adaptive strategy, $\mA_{\ell}$ depends on $\bm{\Phi}_{\ell-1}$. After $L$ layers, the network constructs $\bm{\Phi}_L$. If we denote $\mC_{\vtheta|\bm{\Phi}_L}$ for the covariance of the Bayesian posterior $\rho(\vtheta| \bm{\Phi}_L)$, then it can be shown that the Bayesian minimum mean squared error (MMSE) loss averaged over query tokens for a given $\bm{\Phi}_L$
is $\text{Tr}(\mC_{\vtheta|\bm{\Phi}_L})$ (App.~\ref{app:posterior_mmse_linreg}). The average MMSE loss is $\langle \text{Tr}(\mC_{\vtheta|\bm{\Phi}_L}) \rangle$, where the expectation is over contexts $\mathcal{Z}_N$. Next, we compare the MMSE loss of one-layer and multi-layer architectures while keeping the total communication budget $M = dL$ fixed. We perform this analysis for a Gaussian prior and a novel Tree prior over $\vtheta$. 

\begin{figure}[t]
    \begin{center}
        \includegraphics[width=0.75\linewidth]{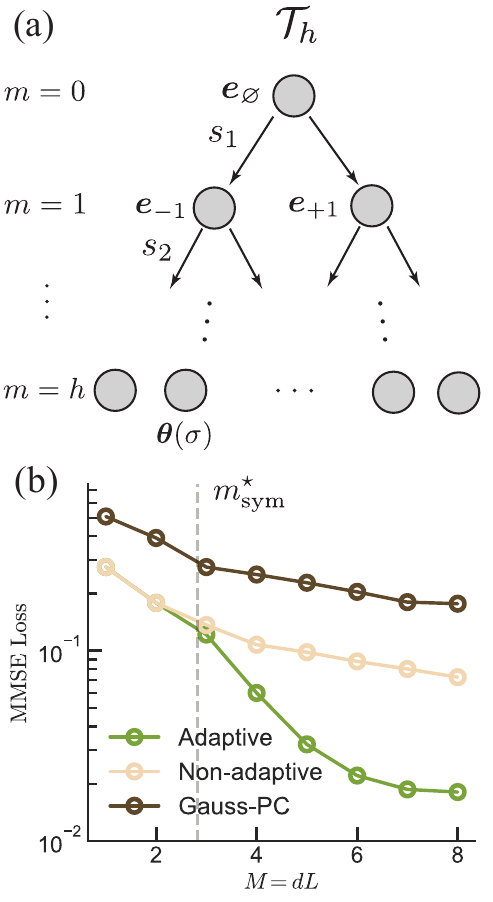}
    \end{center}
    \caption{\textbf{Adaptive and non-adaptive inference strategies for the tree prior.} (a) Each internal node of the depth $h$ tree is a coordinate. The context vectors $\vtheta(\sigma)$ are weighted sums of coordinates along the path to a leaf $\sigma = s_1 s_2 \dots s_h$. (b) The MMSE loss from three strategies. Adaptive: A multi-layer strategy that adaptively infers the path towards the leaf by making a single measurement at every layer. Non-Adaptive: An optimized single-layer strategy with a measurement $\bm{W}$ whose rows are in the span of the depth-symmetric axes and the top $2^4 - 1$ eigenvectors of the prior covariance. Gauss-PC: The expected MMSE loss from a one-layer rank-constrained linear attention transformer without MLP blocks. Here, $ h= 8, \alpha = 0.5, \tau = 0.1$. For these values, $m^{\star}_{\text{sym}} \approx 2.8$ and $m^{\star}_{\text{ada}} \approx 5.6$.}
    \label{fig:tree_prior}
\end{figure}

\subsection{Gaussian Prior}

Consider a Gaussian prior with covariance $\bm{C}_0$, $\vtheta \sim \mathcal{N}(\bm{0}, \bm{C}_0)$, for which we find (details in App.~\ref{app:gaussian_prior_theory})
\begin{equation}
    \mC_{\vtheta|\bm{\Phi}_\ell} = \mC_0 - \mC_0 \mA_{\ell}^\top\left[\mA_{\ell}(\mC_0+\tau^2 \mI_K)\mA_{\ell}^\top\right]^{\dagger}\mA_{\ell}\bm{C}_0.
\end{equation}
Crucially, the expression for $\text{Tr}(\bm{C}_{\vtheta|\bm{\Phi}_L})$ does not depend on the function vector $\bm{\Phi}_L$ for a fixed measurement matrix $\mA_L$. The expected posterior covariance is fully determined by the choices of which measurements the strategy chooses in subsequent layers rather than the observed outcomes of measurements. The optimal adaptive strategy at layer $\ell$ is thus to pick a $\mW_{\ell+1}$ with $d$ rows in the span of the top $d(L - \ell)$ eigenvectors of $\mC_{\vtheta | \bm{\Phi}_{\ell}}$. However, the total MMSE loss from this adaptive strategy is identical to the non-adaptive strategy of choosing $\mA_L$ from the top $dL$ eigenvectors of $\mC_0$ from the outset. Thus, given a communication budget $M$, $L$ layers of interaction dimension $d$ cannot outperform a single layer with interaction dimension $dL$. That a one-layer model is optimal for Gaussian priors was also shown in \citet{bordelon_theory_2026}.

\subsection{Tree Prior}

The property that the posterior covariance does not depend on the measurement is a special feature of Gaussian priors. The theory suggests that depth is useful in the opposite regime where the posterior covariance is strongly informed by the outcome of the measurement. Can one construct such a prior?

Consider a balanced binary tree $\mathcal{T}_h$ of depth $h$ (Figure~\ref{fig:tree_prior}a), where each of the $K = 2^{h}-1$ internal nodes represent orthogonal coordinates. Each leaf $\sigma$ corresponds to a regression vector $\vtheta$, which is specified by the path from the root of the tree to this leaf. Specifically, a leaf $\sigma$, represented as a sequence of spins $s_1s_2\cdots s_h$ ($s_m = \pm 1$), corresponds to an element in the prior:
\begin{equation}
    \vtheta(\sigma)=  \kappa \sum_{m=0}^{h-1} \alpha^{m/2}  s_{m+1}\ve_{s_1s_2\dots s_m}, \label{eq:tree_prior}
\end{equation}
where $0<\alpha<1$, $\kappa$ is a constant that normalizes $\vtheta$, and $\bm{e}_{s_1s_2\dots s_m}$ is the coordinate of internal node $s_1s_2\dots s_m$. The prior is uniform over the $2^h$ leaves of the tree. Given this prior, we would like to predict the MMSE of one-layer and multi-layer models. According to our theory, this MMSE should correspond to the MMSE achieved by the best non-adaptive and adaptive strategies, respectively. 
\\

\noindent \textbf{Non-adaptive strategies.} Consider the non-adaptive strategy of choosing $\mA_L$ as the top $M$ eigenvectors of the prior covariance, as in the Gaussian case. The prior covariance is diagonal, and all coordinates at depth $m$ have variance proportional to $\left(\alpha/2\right)^{m}$. The top eigenvectors are the shallowest tree coordinates. Because the multiplicity of eigenvalues at level $m$ is $2^m$, this strategy resolves spins up to tree depth $m^{\star}_{\text{PC}} \sim \log_2 M$, and the total posterior variance (and thus MMSE loss) scales as $\alpha^{m^{\star}_{\text{PC}}}$, namely, a power of $M$. 

However, it turns out there are more effective non-adaptive strategies; for example, one can measure along axes that are symmetric with respect to the tree coordinates at a particular depth $m$ ($0 \le m \le h-1$),
\begin{equation}
    \vmu_{m}=2^{-m/2}\sum_{v \in \mathcal{V}_m} \ve_v,
\end{equation}
where $\mathcal{V}_m$ are the $2^m$ nodes at depth $m$ and the prefactor is chosen so that $\{\vmu_{m}\}_{m=0}^{h-1}$ are orthonormal. We have  $\vmu_{m}^\top\vtheta(\sigma) = \kappa (\sqrt{\alpha/2})^{m}s_{m+1}$.  If $\mA_L = \vmu_0 \oplus \vmu_1 \oplus \dots \oplus \vmu_{M-1}$ ($M \le h$), then the function vector can be viewed as a noisy representation of $\sigma$ in base $\sqrt{\alpha/2}$. Noise prevents resolving the spin $s_{m+1}$ if $\kappa (\alpha/2)^{m/2} \lesssim \tau$, which means $\sigma$ is resolved up to tree depth $m^{\star}_{\text{sym}} \approx 2\log (\tau/\kappa)/\log (\alpha/2)$. The posterior $\rho(\vtheta|\bm{\Phi}_L)$ is concentrated on leaves that share the prefix until level $m^{\star}_{\text{sym}}$, which implies that the MMSE loss is $\sim  \alpha^{\min(M, m^{\star}_{\text{sym}})}$, i.e., an exponential scaling with $M$ and a cutoff at $m^{\star}_{\text{sym}}$. While this argument shows that there are more effective non-adaptive strategies, it still does not provide the optimal non-adaptive strategy. We instead resort to numerics (described in App.~\ref{app:optimized_nonadaptive}) to compute near-optimal non-adaptive strategies for a given budget $M$. 
\\

\noindent \textbf{Adaptive strategies.} The optimal adaptive strategy is difficult to compute, but a simpler heuristic provides a useful upper bound to the MMSE and is sufficient to explain why depth is useful. Assume $d = 1$ (so $L = M$). The heuristic strategy first measures the root coordinate $\ve_{\varnothing}$. Now, if $\varphi_{1}$ is positive, the next measurement routes to the $+1$ child, and otherwise to the $-1$ child. Repeating this procedure layer by layer, the strategy measures the coordinate of the node selected by the previous signs and routes to the next child according to the sign of that noisy measurement. Thus, the $M$ measurements trace a data-dependent path through the tree until depth $M$. Since $\vtheta$ will be resolved up to depth $M$, the MMSE loss scales as $\alpha^M$. Noise will eventually limit the level to which spins can be resolved. However, importantly, noise now acts only along a single axis. In this case, a spin cannot be resolved when $\kappa \alpha^{m/2} \lesssim \tau$, which leads to a cutoff $m^{\star}_{\text{ada}} \approx 2\log (\tau/\kappa)/ \log \alpha$. Thus, we expect that the MMSE loss still decays exponentially as $\sim \alpha^{\min(M, m_{\text{ada}}^{\star})}$ with a higher cutoff $m_{\text{ada}}^{\star}$ than in the non-adaptive case. 
\\

\noindent \textbf{Computing the MMSE loss.} For the Adaptive and Non-Adaptive strategies, we first sample contexts $\vtheta + \vxi$, obtain $\mA_L,\bm{\Phi}_L$, estimate the posterior $\rho(\vtheta | \bm{\Phi}_L)$ and average $\text{Tr}(\bm{C}_{\vtheta|\bm{\Phi}_L})$ over contexts to obtain the MMSE loss (App.~\ref{app:posterior_mmse_linreg}). For the Non-Adaptive strategy, we take $\mA_{L}= \mU\mD$, $\mU\mU^\top=\mI_M$ and optimize $\mU$. An exhaustive search requires $\mD$ to contain all $2^h - 1$ coordinates, which is computationally challenging; we restrict the rows of $\mA_L$ to the span of $\{\bm{\mu}_{m}\}$ and the top $2^{g} - 1$ eigenvectors of the prior covariance ($g=4$ in numerics).  We call Gauss-PC as the predicted loss from a one-layer linear attention transformer without MLPs (App.~\ref{app:linear-attention-baseline}), which depends on the spectrum of $\mC_0$.  

The MMSE losses for three strategies (labeled Adaptive, Non-Adaptive and Gauss-PC) are plotted in Figure~\ref{fig:tree_prior}b. An adaptive strategy should by construction have MMSE loss at most that of a non-adaptive strategy. Given a communication budget $M$, the heuristic Adaptive strategy splits from the Non-Adaptive strategy for $M >  m^{\star}_{\text{sym}}$ (Figure~\ref{fig:tree_prior}b). Thus, the theory provides a specific prediction that a multi-layer transformer will outperform a one-layer model in this regime, which we now test. 

\section{Adaptive inference in deep linear attention transformers}
\label{sec:tf_results}

\begin{figure*}[t]
    \begin{center}
    \includegraphics[width=\linewidth]{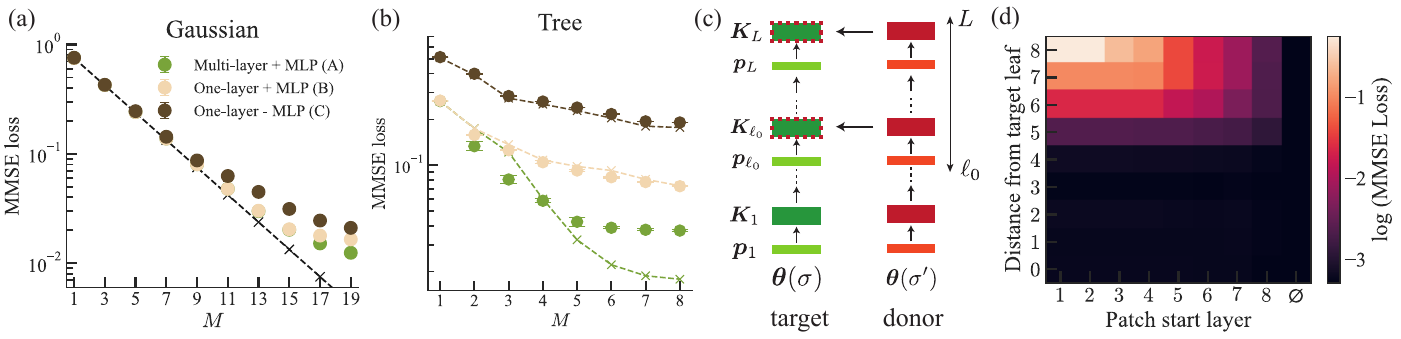}
    \end{center}
    \caption{\textbf{Transformers learn adaptive inference and context-dependent routing.} (a) MMSE loss of Models A--C for a Gaussian prior. (b) MMSE loss for the tree prior. The dashed curves denote the theoretical strategies from Section~\ref{sec:depth}. Error bars denote one standard deviation over training seeds. (c) Key-patching experiment where keys $\mK_\ell$ from a donor leaf ($\sigma'$) are patched into a target leaf ($\sigma$) at every layer $\ell\geq \ell_0$. (d) MMSE loss as a function of donor-to-target tree distance and the intervention layer $\ell_0$. The rightmost column is the no-patch control and the bottommost row is a control where donor contexts are resampled from the target leaf $\sigma$.}
    \label{fig:tf_results}
\end{figure*}

\subsection{Communication-constrained linear attention transformers} 

To test predictions from this model, we consider a communication-constrained $L$-layer linear attention transformer~\citep{wang2020linformer, schlag2021linear}. As discussed previously, the theory applies in a regime where the architecture can retain information about function vectors and the input; we therefore choose a model dimension $D \ge (K+1) + dL$. Denote $S = D - (K+1)$ and the residual stream activation at layer $\ell$ (where $1 \le \ell \le L+1$) and token index $i$ as $\vu_{\ell, i}$. Though not strictly necessary in practice, to maintain interpretability, we explicitly disentangle the input $\vz_i$ and the `internal state' $\vp_{\ell}$ as $\vu_{\ell,i} = \vz_i \oplus \vp_{\ell}$. In the first layer, we initialize $\vp_{1} = \epsilon\bm{1}_S$, where $\epsilon$ is a small constant. Denote $\vphi_{\ell}$ as the local feedforward block (MLP) at layer $\ell$, which takes in $\vp_{\ell}$ as input and outputs a $d \times (K+1)$ matrix $\mK_{\ell}$. $\mQ_{\ell}, \mV_{\ell}, \mO_{\ell}$ are $d \times S$, $d \times (K+1)$ and $S \times d$ matrices. For each layer $\ell$, we iterate
\begin{align}
    \vvarphi_{\ell} &= \frac{1}{N}\sum_{j=1}^N \left(\mV_{\ell}\vz_j\right)\odot \left(\mK_{\ell}\vz_{j}\right), \quad \mK_{\ell} = \bar{\vphi}_{\ell}(\bar{\vp}_{\ell}), \nonumber\\
    \vp_{\ell+1} &= \vp_{\ell} + \frac{1}{\sqrt{d}}\, \mO_{\ell}\left( \vvarphi_{\ell} \odot \mQ_{\ell}\bar{\vp}_{\ell}\right), \label{eq:transformer}
\end{align}
where the bar denotes row/layer normalization and $\odot$ represents elementwise multiplication. Finally, we pass the activation $\vu_{L+1, q}$ at the query token through another feedforward block $\vphi^{\text{out}}$ to generate the output prediction: $\vu_{\text{out}, q} =  \vu_{L+1, q} + \vphi^{\text{out}}(\bar{\bm{u}}_{L+1, q})$. Eq.~\ref{eq:transformer} is a linear attention transformer with three special features compared to common formulations: (1) the elementwise multiplication implements $d$ heads that each convey a single scalar, allowing us to fix the interaction dimension to $d$, (2) the token representation $\vz$ is never updated and is the only information conveyed by the attention heads, and (3) the MLP $\vphi_{\ell}$ reads from the internal state variable $\vp_{\ell}$ and directly affects the key matrix, $\mK_{\ell}$. Under our theoretical framework, we would expect the MLP $\vphi_{\ell}$ to read the internal state $\bar{\vp}_{\ell}$ (where the function vectors from earlier layers are written) and decide on an appropriate measurement, here specified by $\bm{K}_{\ell}$.

\subsection{Depth enables an adaptive search over the hypothesis space}

Predictions from the theory are tested using the architecture in Eq.~\ref{eq:transformer} in three different configurations A, B and C (Figure~\ref{fig:models}, with details in App.~\ref{app:transformer_training}). Model A has a standard arrangement of interleaved attention and MLP blocks with $L=M$ layers and interaction dimension $d=1$. Each $\vphi_{\ell}$ is an MLP and $\vphi^{\text{out}}$ is three stacked MLPs. Model B is a one-layer model with interaction dimension $M$ with identical MLPs as Model A. All MLPs have a single hidden layer with GeLU activation. 
Model C is a standard one-layer linear attention model with interaction dimension $M$ and no feedforward blocks.  It can be shown that Model C implements $\hat{y}(\vx, \mZ_N) = (\mW(\vtheta + \vxi))^\top\vx$, where $\mW$ is a rank-$M$ matrix~\citep{ahn_transformers_2023, zhang2024trained}. Once optimized, $\mW$ projects onto the span of the top $M$ eigenvectors of $\mC_0$ (App.~\ref{app:linear-attention-baseline}).

Figure~\ref{fig:tf_results}a shows the results of training these models on a Gaussian prior $\rho$ with covariance $\mC_0 = \mU\bm{\Lambda} \mU^\top$ where $\bm{\Lambda}$ is diagonal with  $\bm{\Lambda}_{kk} \propto \alpha^k$ (up to a normalization constant) for $k = 1, 2, \dots, K$ and $\mU$ is a random orthogonal matrix (details provided in App.~\ref{app:gaussian_experiments}). Consistent with the theory, all models achieve approximately the same loss: neither MLPs nor depth offer any advantage in this scenario. 

Figure~\ref{fig:tf_results}b shows the results for the tree prior. Consistent with theory, the MMSE loss of Models B and C aligns with the MMSE loss of the Non-Adaptive and Gauss-PC strategies, respectively. The remarkable match of the numerically computed MMSE loss for the Non-Adaptive strategy and the loss of Model B hints that MLP-based one-layer models do achieve optimality. Importantly, Model A outcompetes the optimized Non-Adaptive strategy, showing that the model indeed executes an adaptive inference strategy over its layers. However, the difference between the MMSE loss for Model A and the Adaptive strategy suggests that the network does not achieve optimality. We verified that this gap is not due to finite $N$ effects or an insufficiently expressive MLP ($\vphi_{\ell}$). Thus, while deep models do appear to implement an adaptive strategy, there is further room for improvement. It is an interesting question whether an alternative architecture can achieve better performance under similar constraints. We have also confirmed that the advantages of depth qualitatively persist in conventional linear and softmax self-attention architectures (App.~\ref{app:comparison_tf}).

\subsection{Causal evidence for adaptive routing}

As a direct causal test of the adaptive routing mechanism, we performed a key-patching experiment on Model A ($M = 8,h=8$). If the model performs adaptive inference, the embeddings the model chooses for a context sampled from a leaf $\sigma'$ should fare worse when used to infer the context variable of another leaf $\sigma$, and this difference should depend on the tree distance between $\sigma$ and $\sigma'$. To test this hypothesis, for contexts sampled from the target leaf $\sigma$ and the donor leaf $\sigma'$ at distance $d$ (from $0$ to $h$), we first ran a forward pass of the donor context and recorded the dynamic key matrices $\{\mK^{\sigma'}_{\ell}\}_{\ell=1}^{M}$. We then reran the target context from $\sigma$, but from patch start layer $\ell_0$ onward replaced $\mK^{\sigma}_{\ell}$ by $\mK^{\sigma'}_{\ell}$ for all $\ell\geq \ell_0$, and measured the MMSE loss (Figure~\ref{fig:tf_results}c). A model that does not implement an adaptive strategy should show no or little change in the MMSE loss across perturbations. Figure~\ref{fig:tf_results}d shows that donor keys whose path diverges early from the target produce a large loss when inserted early. The effect is absent for donor leaves that are proximal to the target and for late or no interventions. The rightmost column $\ell_0=\varnothing$ is a no-perturbation control. This experiment was also repeated on a conventional linear self-attention transformer (App.~\ref{app:comparison_tf}) to confirm that it does not specifically rely on our communication-constrained architecture.

\section{Discussion}

A prominent view is that in-context learning emulates Bayesian inference over a latent context variable~\citep{muller2021transformers, xie_explanation_2022}. Mechanistic studies of in-context linear regression (for example) have shown how attention circuits can implement algorithms such as preconditioned gradient descent~\citep{ahn_transformers_2023, oswald_transformers_2023, zhang2024trained}. We make progress towards bridging the gap between these two pictures by showing that deep transformers implement inference as a distributed, adaptive, layerwise procedure. Our framework suggests that function vectors act as compressed state variables, MLP blocks choose which statistic should be measured next, and attention pools that statistic across the context. This mechanism places function vectors~\citep{todd_function_2023, hendel_context_2023, hojel_finding_2024, yin2025attention, gibson_distinct_2026} in a circuit-level framework and shows that depth and MLP blocks enable a much richer class of in-context learning algorithms than preconditioned gradient descent or copying. 

The theory ignores a key element of typical transformer architectures, softmax attention, which allows tokens to retrieve information from specific tokens in the context at the expense of a quadratic scaling of computation with context size~\citep{miller2016key}. Softmax attention is analogous to an implementation of a key-value memory over the context~\citep{bricken2021attention, smart2025context}, and plays an important role in implementing attention-based circuit computations for in-context learning~\citep{olsson_-context_2022, bietti2023birth, reddy2023mechanistic, gibson_distinct_2026}. The pooling-based operation in our theory is linear in context size while respecting exchangeability~\citep{diaconis1980finite, zaheer2017deep, edwards2016towards}, namely, permutation invariance of the data conditional on the latent context variable, and thus may be sufficient to implement in-context learning in such scenarios. 

Recent analyses that treat self-attention as an interacting-particle system similarly emphasize a collective description of token dynamics, but focus on the evolution of the empirical distribution of token representations and the emergence of clustering or metastable macroscopic states~\citep{geshkovski2023emergence, bruno2025emergence, sander2022sinkformers}. In a complementary direction, in-context denoising identifies regimes in which attention implements a context-dependent associative-memory or score-like denoising update~\citep{smart2025context,rosu2026softmax}. Integrating these perspectives with the constrained Bayesian framework developed here may clarify how transformers combine local memory retrieval with adaptive inference using feedforward blocks. 

\subsection*{AI use statement}

In this work, we used generative AI tools (OpenAI Codex, GPT-5.5) for coding assistance in writing experiments. We did not use generative AI tools for other any tasks requiring disclosure, including the development of theoretical models or conceptual frameworks, formulation or proof of mathematical claims, development or refinement of hypotheses, design of research methodology or experiments, or interpretation of results. Other required-disclosure tasks, including translation, dataset cleaning or reformatting, and qualitative or thematic data analysis, are not applicable to this work. Generative AI was used to check for grammatical and typesetting errors, but was not used to generate scientific results, theoretical claims, writing, or interpretations.

\subsection*{Reproducibility statement}

All code used to generate the results and figures of this work is available in the supplementary code files at \texttt{icl\_function\_vectors.zip}.

\subsubsection*{Acknowledgments}
We thank Emmy Blumenthal, Nikolas Claussen, Cole Gibson, and Albert Qin for helpful comments on the manuscript, and Adel Ardalan, Tim Buschman, Declan Campbell, Jonathan Cohen and Tim Lin for insightful discussions. GR was partially supported by a joint research agreement between NTT Research Inc. and Princeton University, a grant from Coefficient Giving and a seed grant from the Princeton AI Lab. The simulations presented in this article were performed on computational resources managed and supported by Princeton Research Computing, a consortium of groups including the Princeton Institute for Computational Science and Engineering (PICSciE) and Research Computing at Princeton University.

\bibliography{references}
\bibliographystyle{iclr2027_conference}

\appendix
\section{Experimental Details}

\subsection{Transformer Architecture and Training}
\label{app:transformer_training}

We optimize all transformer models using AdamW with batch size $128$, learning rate $3\times10^{-4}$, and weight decay $10^{-4}$. The model dimension is $D=2(K+1)$, where $K$ is the dimensionality of $\vx$ and $\vtheta$.

All feedforward blocks use a single hidden layer with Gaussian Error Linear Unit (GeLU) activations and hidden dimension $D$. The output block $\vphi_{\mathrm{out}}$ consists of three stacked MLPs followed by a linear projection to a scalar output, which gives the prediction $\hat{y}(\vx,\mathcal{Z}_N)$ for the query token.

We write the residual stream as $\vu_{\ell,i}=\vz_i\oplus\vp_\ell$, where
$\vz_i\in\mathbb{R}^{K+1}$ contains the fixed token representation and $\vp_\ell\in\mathbb{R}^S$ is the shared internal state, with $S=D-(K+1)$. The internal state is initialized as
\begin{equation}
    \vp_1=\epsilon\mathbf{1}_S,
\end{equation}
where $\epsilon$ is a small positive constant. For a model with interaction dimension $d$, the matrices $\mQ_\ell$, $\mK_\ell$, $\mV_\ell$, and $\mO_\ell$ have dimensions $d\times S$, $d\times(K+1)$, $d\times(K+1)$, and $S\times d$, respectively. In the implementation, the internal state supplied to the MLP and query branches is layer-normalized. Thus, the dynamic key matrix is constructed from the normalized state.

Query tokens are embedded in the same residual-stream representation as context tokens, except that their output component is masked. In particular, for a query token we set $\vz=(\vx,0)$. For all transformer models trained in Sec.~\ref{sec:tf_results}, the block diagrams of the models are shown in Fig.~\ref{fig:models}. All other architectural details follow the communication-constrained transformer defined in Appendix~\ref{app:transformer_mapping}.

\begin{figure}[t]
    \centering
    \includegraphics[width=0.8\linewidth]{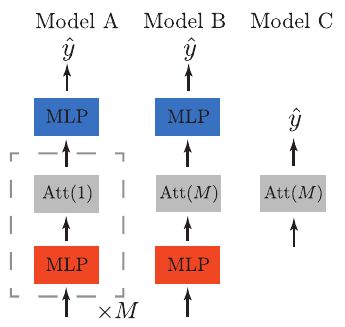}
    \caption{Three transformer configurations. A: $M$-layer model with interaction dimension 1 per layer with an MLP embedding function (orange) and an MLP decoder (blue). B: 1-layer model with interaction dimension $M$ with an MLP embedding function and MLP decoder. C: 1-layer with interaction dimension $M$ and no MLPs.}
    \label{fig:models} 
\end{figure}

\subsection{Gaussian Prior Experiments}
\label{app:gaussian_experiments}

For the Gaussian-prior experiments, we sample
\begin{equation}
    \vtheta\sim\mathcal{N}(0,\mC_0),\qquad \mC_0=\mU\mLambda\mU^\top,
\end{equation}
where $\mU$ is a random orthogonal matrix and $\mLambda$ is diagonal with
\begin{equation}
    \Lambda_{kk}=\frac{1-\alpha}{\alpha(1-\alpha^K)}\alpha^k, \qquad k=1,\ldots,K.
\end{equation}
For the results shown in Figure~\ref{fig:tf_results}, we use $\alpha=0.75$ and $\tau=0.01$.

Each context contains $N=4096$ context tokens, and the prediction loss for each context is evaluated using $256$ query tokens. We use a relatively large value of $N$ in these experiments to suppress finite-$N$ effects, which otherwise introduce a lower bound to the achievable MMSE at larger communication budgets.

We train $8$ independent seeds for each model configuration for $20000$ iterations.

\subsection{Tree Prior Experiments}
\label{app:tree_prior_experiments}

For all tree-prior experiments, we use $h=8$, $\alpha=0.5$, and $\tau=0.1$. For the trained transformer results in Figure~\ref{fig:tf_results}, each context contains $N=512$ context tokens, and the prediction loss is evaluated using $64$ query tokens. We train $8$ independent seeds for each model configuration for $40000$ iterations.

To test whether the gap between the learned deep model and the heuristic adaptive strategy is caused by finite-context effects or insufficient model capacity, we repeated the experiments with $N=1024$ and with additional layers in the layerwise MLPs. Neither modification produced a meaningful change in the MMSE loss.

For the theoretical strategy curves in Figure~\ref{fig:tree_prior}, the prior contains $2^h$ regression vectors in a $K=2^h-1$-dimensional space. For each measurement strategy, we estimate the expected MMSE by averaging the posterior variance over contexts and measurement noise. For a row-orthonormal measurement matrix $\mW$ satisfying $\mW\mW^\top=\mI_M$, the function vector is
\begin{equation}
    \vPhi=\mW(\vtheta(\sigma)+\vxi),\qquad
    \vxi\sim\mathcal{N}(0,\tau^2\mI_K).
\end{equation}
For a fixed $\mW$, the posterior probability of leaf $\sigma$ is
\begin{equation}
    \rho_{\mW}(\sigma|\vPhi)=\frac{\exp\left[-\|\vPhi-\mW\vtheta(\sigma)\|^2/(2\tau^2)\right]}{
    \sum_{\sigma'}\exp\left[-\|\vPhi-\mW\vtheta(\sigma')\|^2/(2\tau^2)\right]}.
\end{equation}
The corresponding posterior mean is
\begin{equation}
    \hat{\vtheta}(\vPhi)=\sum_\sigma\vtheta(\sigma)\rho_{\mW}(\sigma|\vPhi),
\end{equation}
and the posterior MMSE is
\begin{equation}
    \Tr
    \left(\mC_{\vtheta|\vPhi}\right)= \sum_\sigma\left\|\vtheta(\sigma)-\hat{\vtheta}(\vPhi)\right\|^2\rho_{\mW}(\sigma|\vPhi).
\end{equation}
We enumerate all $2^h$ leaves exactly when evaluating the posterior and average this quantity over samples of the measurement noise. Since every leaf has unit norm, the posterior MMSE may equivalently be written as
\begin{equation}
    \Tr\left(\mC_{\vtheta|\vPhi}\right)=1-\left\|\hat{\vtheta}(\vPhi)\right\|^2.
\end{equation}

\subsubsection{Optimized non-adaptive baseline}
\label{app:optimized_nonadaptive}

An exhaustive optimization over all $M$-dimensional subspaces of the $K=2^h-1$ tree coordinates is computationally challenging. We therefore optimize over a restricted subspace that includes both the depth-symmetric directions and individual coordinates from the shallow levels of the tree. Specifically, define
\begin{equation}
    \mathcal{H}=\operatorname{span}
    \left(
    \{\ve_v:\operatorname{depth}(v)<g\}\cup\{\vmu_m\}_{m=0}^{h-1}
    \right).
\end{equation}
The individual tree coordinates in the first set correspond to the top $2^g-1$ eigenvectors of the prior covariance. We use $g=4$ for the
results in Figure~\ref{fig:tree_prior}. Let $\mD$ be a row-orthonormal basis for $\mathcal{H}$. We parameterize the measurement matrix as
\begin{equation}
    \mW=\mU\mD,\qquad\mU\mU^\top=\mI_M.
\end{equation}
For each $\mU$, we enumerate all leaves exactly and evaluate the posterior defined above. Since $\|\vtheta(\sigma)\|=1$, minimizing the expected MMSE is equivalent to maximizing
\begin{equation}
    J(\mU)=\left\langle
    \left\|\sum_\sigma\vtheta(\sigma)\rho_{\mU\mD}\left(\sigma|\mU\mD(\vtheta(\sigma_0)+\vxi)\right)\right\|^2
    \right\rangle_{\sigma_0,\vxi}.
\end{equation}

We optimize $\mU$ using simultaneous perturbation stochastic approximation (SPSA) subject to the row-orthonormality constraint $\mU\mU^\top=\mI_M$~\citep{119632}. At each optimization step, for a random sign matrix $\mDelta$, we evaluate $J$ at two re-orthonormalized perturbations of $\mU$ and use their finite difference to construct a gradient estimate $\mG$. The gradient is projected onto the tangent space of the row-orthonormal constraint using
\begin{equation}
    \Pi_{\mU}(\mG)=\mG-\operatorname{Sym}(\mG\mU^\top)\mU,
\end{equation}
where $\operatorname{Sym}(\mA)=\frac{1}{2}\left(\mA+\mA^\top\right)$. We then update $\mU$ along this projected direction and re-orthonormalize its rows. The SPSA step size and perturbation schedules are $a_k=\eta k^{-0.602}$ and $c_k=\epsilon k^{-0.101}$ respectively.

\subsubsection{Adaptive strategy}

For the heuristic adaptive strategy, we take $d=1$ and simulate the measurements sequentially. The first measurement is taken along $\mathbf{e}_{\varnothing}$. At every subsequent step, the sign of the previous measurement determines which child coordinate is measured next. If
\begin{equation}
    \hat{\sigma}=\hat{s}_1\hat{s}_2\cdots\hat{s}_M
\end{equation}
denotes the path selected by these noisy measurements, the realized measurement directions are
\begin{equation}
    \mW=\ve_{\varnothing}\oplus\ve_{\hat{s}_1}\oplus\cdots\oplus\ve_{\hat{s}_1\cdots\hat{s}_{M-1}}.
\end{equation}
For each sampled noise realization, we construct the resulting function vector, enumerate the posterior over all leaves, and compute the posterior MMSE. By symmetry of the tree prior, the noise average can be evaluated using contexts generated from a single fixed leaf.

\subsubsection{Gauss-PC baseline}

The Gauss-PC baseline corresponds to the optimal rank-$M$ linear estimator obtained from the prior covariance alone. If
$\lambda_1\geq\lambda_2\geq\cdots$ are the eigenvalues of $\mC_0$, its expected squared prediction loss is
\begin{equation}
    \mathcal{L}_{\mathrm{Gauss-PC}} =\Tr(\mC_0)-\sum_{i=1}^M\frac{\lambda_i^2}{\lambda_i+\tau^2}.
\end{equation}
This baseline selects the top $M$ eigenvectors of the prior covariance, as would be optimal for a Gaussian prior, but does not exploit the non-Gaussian hierarchical structure of the tree prior.

\section{Theoretical Details}

\subsection{Bayesian Posterior MMSE for In-Context Linear Regression}
\label{app:posterior_mmse_linreg}

For the linear regression problem considered in the main text, the posterior mean estimator of the regression vector given a function vector $\vPhi$ is
\begin{equation}
    \hat{\vtheta}(\vPhi) = \int d\vtheta\, \vtheta\, \rho(\vtheta|\vPhi).
\end{equation}
The corresponding Bayes-optimal prediction for a query $\vx$ is
\begin{equation}
    \hat{y}(\vx,\vPhi) = \hat{\vtheta}(\vPhi)^\top\vx.
\end{equation}
Conditioned on $\vPhi$, the query-averaged squared prediction error is
\begin{align}
    &\left\langle
    \left(
    \vtheta^\top\vx -\hat{y}(\vx,\vPhi)
    \right)^2
    \right\rangle_{\vx,\vtheta|\vPhi} \nonumber \\
    &\qquad= \left\langle
    \left[
    (\vtheta-\hat{\vtheta}(\vPhi))^\top\vx
    \right]^2
    \right\rangle_{\vx,\vtheta|\vPhi} \\
    &\qquad=\left\langle
    \vx^\top
    (\vtheta-\hat{\vtheta}(\vPhi))(\vtheta-\hat{\vtheta}(\vPhi))^\top \vx\right\rangle_{\vx,\vtheta|\vPhi}.
\end{align}
Using $\langle\vx\vx^\top\rangle=\mI_K$, we thus obtain
\begin{equation}
    \left\langle
    \left(\vtheta^\top\vx-\hat{y}(\vx,\vPhi)\right)^2
    \right\rangle_{\vx,\vtheta|\vPhi}
    =\Tr\left(\mC_{\vtheta|\vPhi}\right),
\end{equation}
where
\begin{equation}
    \mC_{\vtheta|\vPhi}=\left\langle
    (\vtheta-\hat{\vtheta}(\vPhi))(\vtheta-\hat{\vtheta}(\vPhi))^\top
    \right\rangle_{\vtheta|\vPhi}
\end{equation}
is the posterior covariance. Thus, minimizing the expected prediction error is equivalent to minimizing the expected trace of the posterior covariance.

\subsection{Gaussian Prior}
\label{app:gaussian_prior_theory}

Consider the Gaussian prior $\vtheta\sim\mathcal{N}(0,C_0)$, and suppose that after $\ell$ layers the accumulated function vector takes the form
\begin{equation}
    \vPhi_\ell=\mA_\ell(\vtheta+\vxi),\qquad \vxi\sim\mathcal{N}(0,\tau^2\mI_K),
\end{equation}
where
\begin{equation}
    \mA_\ell=\mW_1\oplus\mW_2\oplus\cdots\oplus\mW_\ell.
\end{equation}
For a fixed choice of $\mA_\ell$, $\vtheta$ and $\vPhi_\ell$ are jointly Gaussian, and therefore the posterior $\rho(\vtheta|\vPhi_\ell)$ is also Gaussian. Its covariance is
\begin{equation}
    \mC_{\vtheta|\vPhi_\ell} = \mC_0-\operatorname{Cov}(\vtheta,\vPhi_\ell)\operatorname{Cov}(\vPhi_\ell,\vPhi_\ell)^\dagger\operatorname{Cov}(\vPhi_\ell,\vtheta).
\end{equation}
The required covariance matrices are
\begin{align}
    \operatorname{Cov}(\vtheta,\vPhi_\ell)&= \mC_0\mA_\ell^\top, \\
    \operatorname{Cov}(\vPhi_\ell,\vPhi_\ell)&= \mA_\ell(\mC_0+\tau^2\mI_K)\mA_\ell^\top.
\end{align}
Substituting these expressions gives
\begin{equation}
    \mC_{\vtheta|\vPhi_\ell}= \mC_0-\mC_0\mA_\ell^\top\left[
    \mA_\ell(\mC_0+\tau^2\mI_K)\mA_\ell^\top
    \right]^\dagger\mA_\ell \mC_0,
\end{equation}
which is exactly the result used in the main text.

\subsection{One-Layer Linear Attention Baseline}
\label{app:linear-attention-baseline}

We next derive the optimal prediction of the one-layer linear attention model without MLP blocks, corresponding to model C in Section~\ref{sec:tf_results}. For a context summarized by $\vtheta+\vxi$, where $\vxi\sim\mathcal{N}(0,\tau^2\mI_K)$, the model prediction can be written as
\begin{equation}
    \hat{y}(\vx;\vtheta+\vxi)= \left[\mK(\vtheta+\vxi)\right]^\top\mQ\vx,
\end{equation}
where $\mQ$ and $\mK$ have rank at most $M$. Defining $\mB=\mQ^\top\mK$, the query-averaged prediction loss is
\begin{align}
    \mathcal{L}_{C}(\mB)&=
    \left\langle
    \left\|\vtheta-\mB(\vtheta+\vxi)\right\|^2
    \right\rangle_{\vtheta,\vxi} \\
    &=\Tr\left[(\mI_K-\mB)^\top
    (\mI_K-\mB)\mC_0\right]+\tau^2\|\mB\|_F^2,
\end{align}
subject to $\operatorname{rank}(\mB)\leq M$. Let
\begin{equation}
    \mC_0=\sum_{i=1}^K\lambda_i\ve_i\ve_i^\top,\qquad
    \lambda_1\geq\lambda_2\geq\cdots\geq\lambda_K.
\end{equation}
Optimizing over rank-$M$ matrices gives
\begin{equation}
    \mB^\star=\sum_{i=1}^M\frac{\lambda_i}{\lambda_i+\tau^2}\ve_i\ve_i^\top.
\end{equation}
Thus, the optimal one-layer linear attention model measures the top $M$
eigenvectors of the prior covariance, with noise-dependent shrinkage. Its minimum achievable loss is
\begin{equation}
    \mathcal{L}_{C}^\star=\Tr(\mC_0)-\sum_{i=1}^M\frac{\lambda_i^2}{\lambda_i+\tau^2}.
\end{equation}
This is the \textit{Gauss-PC} prediction used throughout the main text.

\subsection{Tree Prior}

For the tree prior introduced in Section~\ref{sec:depth}, the normalization constant is
\begin{equation}
    \kappa=\sqrt{\frac{1-\alpha}{1-\alpha^h}},
\end{equation}
such that $\|\vtheta(\sigma)\|=1$ for every leaf $\sigma$.

We next derive the spectrum of the prior covariance. A node at depth $m$ lies on the path to a uniformly sampled leaf with probability $2^{-m}$. Conditional on lying on the path, the corresponding component of $\vtheta(\sigma)$ has magnitude $\kappa\alpha^{m/2}$. The variance of each coordinate at depth $m$ is therefore
\begin{equation}
    \kappa^2\left(\frac{\alpha}{2}\right)^m.
\end{equation}
Since there are $2^m$ coordinates at depth $m$, the prior covariance has a $2^m$-fold degenerate eigenvalue proportional to $(\alpha/2)^m$ at each level, with eigenvalues decreasing from the shallowest to the deepest levels of the tree.

If the first $m$ branch choices of a leaf are resolved exactly, the remaining posterior uncertainty is contained in the coordinates below depth $m$. Its total variance is
\begin{equation}
    \kappa^2\sum_{r=m}^{h-1}\alpha^r =\frac{\alpha^m(1-\alpha^{h-m})}{1-\alpha^h} \sim\alpha^m.
\end{equation}
This gives the MMSE scalings used in Section~\ref{sec:depth}.

\subsection{Relation to Linear Attention Transformers}
\label{app:transformer_mapping}

Here we show how the communication-constrained model introduced in Section~\ref{sec:problem_setup} can be implemented by a linear attention transformer. Consider a residual stream activation $\vu_{\ell,i}\in\mathbb{R}^D$ for token $i$ at layer $\ell$. A standard linear attention layer takes the form
\begin{equation}
    \vu_{\ell+1,i} = \vu_{\ell,i} + \mO_\ell
    \left[\left(
    \frac{1}{N}\sum_{j=1}^N (\mV_\ell\vu_{\ell,j})(\mK_\ell\vu_{\ell,j})^\top
    \right) \mQ_\ell\vu_{\ell,i}\right],
\end{equation}
where $\mQ_\ell$, $\mK_\ell$, $\mV_\ell$, and $\mO_\ell$ are the learned query, key, value, and output matrices at layer $\ell$ respectively.

To restrict the interaction dimension to $d$, we consider $d$ independent attention heads with one-dimensional query and key representations. Each head then communicates a single scalar statistic, and the attention update can then be written as
\begin{align}
    \vu_{\ell+1,i}&= \vu_{\ell,i} + \mO_\ell \left[\vvarphi_\ell\odot (\mQ_\ell\vu_{\ell,i})\right], \\
    \vvarphi_\ell&=\frac{1}{N}\sum_{j=1}^N(\mV_\ell \vu_{\ell,j})\odot (\mK_\ell\vu_{\ell,j}),
\end{align}
where $\odot$ denotes elementwise multiplication. The vector $\vvarphi_\ell\in\mathbb{R}^d$ is thus a pooled statistic of the context, with each attention head contributing a single scalar communication channel. This has the same form as the pooled statistic introduced in Sec.~\ref{sec:problem_setup}.

To allow the statistic communicated at each layer to depend on information acquired in previous layers, we partition the residual stream as $\vu_{\ell,i}= \vz_i\oplus \vp_\ell$ where $\vz_i$ stores the original token and $\vp_\ell$ is a shared internal state that evolves across layers. We also take $\vp_\ell$ to have sufficient dimensionality to retain the statistics accumulated across previous layers. We constrain the attention operation so that only the fixed token representation $\vz_i$ is communicated between tokens, while $\vp_\ell$ controls how that information is processed. In particular, the feedforward block (MLP) at layer $\ell$ uses the current internal state to determine the key matrix $\mK_\ell = \vphi_\ell(\vp_\ell)$. Under this restriction, the pooled statistic defined above becomes
\begin{equation}
    \vvarphi_\ell = \frac{1}{N}\sum_{j=1}^N (\mV_\ell \vz_j)\odot (\mK_\ell \vz_j).
\end{equation}
The pooled statistic is then written back into the shared internal state through
\begin{equation}
    \vp_{\ell+1}=\vp_\ell + \frac{1}{\sqrt{d}}\mO_\ell\left[
    \vvarphi_\ell \odot (\mQ_\ell \bar{\vp}_\ell)\right],
\end{equation}
where a factor of $1/\sqrt{d}$ is included as a form of width normalization for the residual update, ensuring that the magnitude of the communicated signal remains $\mathcal{O}(1)$ as the interaction dimension $d$ is varied. Thus, $\vp_{\ell+1}$ contains information acquired from the context at the current layer and can in turn determine the key matrix $\mK_{\ell+1}$ used at the next layer.

\section{Experimental Comparisons Against Standard Architectures}
\label{app:comparison_tf}

To test whether the results in Fig.~\ref{fig:tf_results} depend on the communication-constrained linear attention transformer used in our main experiments, we repeated these experiments using conventional (unnormalized) linear self-attention (LSA) and softmax self-attention (SSA).

\begin{figure*}[t]
    \begin{center}
    \includegraphics[width=\linewidth]{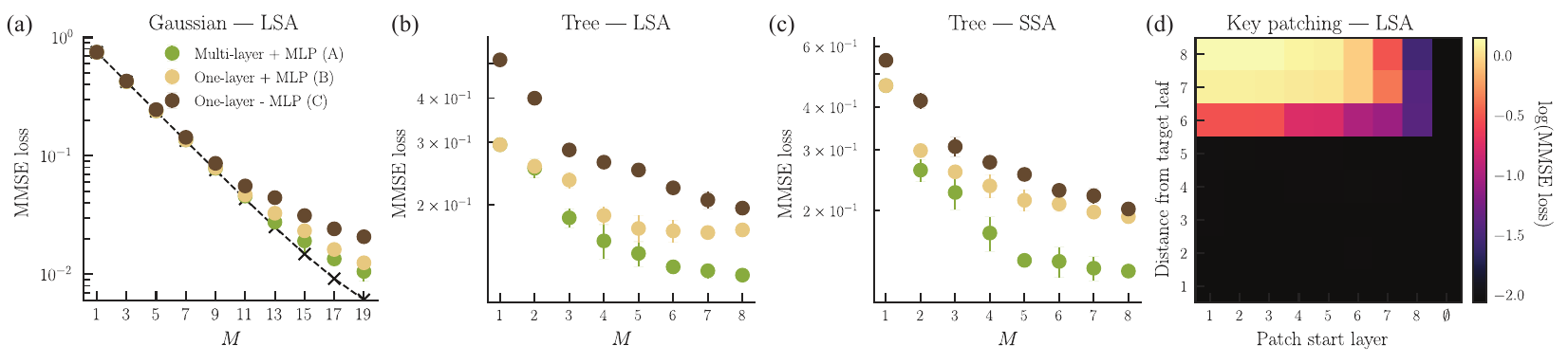}
    \end{center}
    \caption{\textbf{Conventional Architectures Exhibit Depth-Based Advantages.} (a) MMSE loss of Models A--C analogues for linear self-attention (LSA) on the Gaussian prior. The dashed line denotes the theoretical floor from Section~\ref{sec:depth}. (b) MMSE loss of Models A--C analogues for LSA on the tree prior. (c) MMSE loss of Models A--C analogues for SSA on the tree prior. Error bars denote one standard deviation over 8 training seeds. (d) MMSE loss as a function of donor-to-target tree distance and the intervention layer $\ell_0$, for optimized LSA transformers. The rightmost column is the no-patch control and the bottommost row is a control where donor contexts are resampled from the target leaf $\sigma$.}
    \label{fig:comparison_results}
\end{figure*}

\subsection{Linear Self-Attention}

Context tokens were initialized as
\begin{equation}
    \vu_{1,j}=\left(\vx_j,y_j,K^{-1}\mathbf{1}_{K+1}\right),
    \qquad j=1,\ldots,N,
\end{equation}
while query tokens are initialized as
\begin{equation}
    \vu_{1,q}=\left(\vx_q,0,K^{-1}\mathbf{1}_{K+1}\right),
\end{equation}
where the target is once again masked out. Every attention head is restricted to have key and value dimensions of $d_k=d_v=1$. For head $h$ in layer $\ell$, we compute
\begin{equation}
    q_{\ell,h,i}= \mW^Q_{\ell,h}\bar{\vu}_{\ell,i},\quad
    k_{\ell,h,j}= \mW^K_{\ell,h}\bar{\vu}_{\ell,j},\quad
    v_{\ell,h,j}= \mW^V_{\ell,h}\bar{\vu}_{\ell,j},
\end{equation}
where all three quantities are scalars, and the bars denote layer normalization. The un-normalized linear attention output is then
\begin{equation}
    a_{\ell,h,i}=\frac{1}{N}\sum_{j=1}^{N}q_{\ell,h,i}k_{\ell,h,j}v_{\ell,h,j}=
    q_{\ell,h,i}\left(\frac{1}{N}\sum_{j=1}^{N}k_{\ell,h,j}v_{\ell,h,j}\right).
    \label{eq:standard_linear_attention}
\end{equation}
Both context and query tokens are allowed to attend to the $N$ context tokens, while the query tokens do not attend to themselves or each other. For $H_\ell$ heads, the residual attention update is then computed as
\begin{equation}
    \widetilde{\vu}_{\ell,i}=\vu_{\ell,i}+\frac{1}{\sqrt{H_\ell}}\,\mW^O_\ell
    \left(a_{\ell,1,i},\ldots,a_{\ell,H_\ell,i}\right)^\top,
\end{equation}
where the linear attention outputs are concatenated across the $H_\ell$ heads before output projection by $\mW^O_\ell$. Note that the usual $1/\sqrt{d_{\mathrm{head}}}$ dot-product scaling resolves to unity here since $d_{\mathrm{head}}=1$. When present, the feedforward blocks are two-layer pre-LN residual MLPs with GeLU activations and hidden dimension $D$.

We considered the same three communication budget configurations as in the main experiments of Fig.~\ref{fig:tf_results}. Model A used $L=M$ layers with one scalar attention head and one MLP per layer. Model B used a single layer with $M$ scalar attention heads and one MLP. Model C used a single layer with $M$ scalar attention heads and no feedforward blocks. Models A and B were also followed by three additional residual MLP blocks. Thus, each architecture was restricted to $M$ scalar attention channels in total.

For the Gaussian prior experiments, we used the same data distribution, model dimension, and context settings as described in App.~\ref{app:gaussian_experiments}, while for the tree prior experiments, we used the same settings as described in App.~\ref{app:tree_prior_experiments}. For optimization in both cases, models were trained using AdamW for $4000$ steps with batch size $128$, learning rate $3\times10^{-4}$, weight decay $10^{-4}$, and gradient-norm clipping at one.

\subsection{Softmax Self-Attention}

We additionally replaced the un-normalized linear self-attention operation with conventional scaled dot-product softmax attention, while leaving the residual stream, scalar-head restriction, feedforward blocks, and model structures otherwise unchanged.

Using the queries, keys, and values defined above, the output of head $h$ was
\begin{equation}
    a_{\ell,h,i}=\sum_{j=1}^{N} \frac{\exp\!\left(q_{\ell,h,i}k_{\ell,h,j}/\sqrt{d_{\mathrm{head}}}\right)}{\sum_{r=1}^{N}\exp \left(q_{\ell,h,i}k_{\ell,h,r}/\sqrt{d_{\mathrm{head}}}\right)}v_{\ell,h,j}.
\end{equation}
All heads were still constrained to be scalar, keeping $d_k=d_v=1$. As before, context and query tokens attended to all context tokens, while queries did not attend to themselves or to other queries. The resulting head outputs were combined using the same output projection and residual update as in the linear-attention models. Although the number of scalar attention heads is kept fixed, it should be noted that the softmax operation does not impose the same global communication constraint that linear self-attention does. Specifically, each softmax attention head can retrieve a different scalar summary for each query, while each linear attention head computes a single context-pooled scalar shared across all context tokens.

The tree prior experiments were conducted in an identical fashion to those described in App.~\ref{app:tree_prior_experiments}. For optimization, models were trained using AdamW for $4000$ steps with batch size $128$, learning rate $3\times10^{-4}$, weight decay $10^{-4}$, and gradient-norm clipping at one.

\subsection{Results}

We first demonstrated the similarity of conventional un-normalized LSA transformers on the Gaussian prior for all three models (Fig.~\ref{fig:comparison_results}(a)), and confirm that the results are once again consistent with the theory: neither depth nor MLPs provide any advantage in this scenario.

Next, we studied the behavior of optimized LSA and SSA transformers in the tree prior setting, and recover qualitatively similar results to Fig.~\ref{fig:tf_results}(b), as shown in Fig.~\ref{fig:comparison_results}(b) and (c) respectively. Specifically, the scale of the losses are higher than before, but we still observe a similar splitting in the performance of Models A and B at $m=m_\mathrm{sym}^\ast$, confirming that depth still provides an advantage in the conventional architectures being considered for this hierarchically structured prior.

Finally, we confirmed that these intermediate representations are present in the LSA transformer by performing a similar causal key-patching experiment to Fig.~\ref{fig:tf_results}(d), as seen in Fig.~\ref{fig:comparison_results}(d). Since there is no dynamically generated key matrix $\mK_\ell$ here, we instead perform the key patching by first running a forward pass on the tokens for the donor leaf. We then record the generated keys for all tokens, and patch these for all layers $\ell>=\ell_0$ on the target's forward pass, where the tokens for the target's forward pass are generated using the same covariates as that of the donor's forward pass. In a similar vein to the main experiments, we see that donor keys at a greater tree distance from the target produce larger losses. The effect is once again absent for donor leaves that are close to the target and for interventions carried out in later layers.

\end{document}